\documentclass[conference,]{IEEEtran}
\IEEEoverridecommandlockouts

\usepackage{hhline}
\usepackage[numbers]{natbib}
\usepackage[utf8]{inputenc}
\usepackage{graphicx}
\usepackage{amsfonts}
\usepackage{booktabs}
\usepackage{multirow}
\usepackage{threeparttable}
\usepackage{eurosym,makecell,comment}
\usepackage{url}
\usepackage{xcolor}
\usepackage{tikz}
\usepackage{booktabs}
\usepackage{float}
\usepackage{amsmath}
\usepackage{color, colortbl}
	
\definecolor{Gray}{gray}{0.9}
\definecolor{LightCyan}{rgb}{0.88,1,1}

\usepackage{array}
\newcolumntype{P}[1]{>{\raggedright\arraybackslash}p{#1}}
\usepackage{dblfloatfix}

\DeclareRobustCommand*{\IEEEauthorrefmark}[1]{%
  \raisebox{0pt}[0pt][0pt]{\textsuperscript{\footnotesize #1}}%
}

\begin{document}

\title{ProFe: Communication-Efficient Decentralized Federated Learning via Distillation and Prototypes}

\author{
    \IEEEauthorblockN{Pedro Miguel S\'anchez S\'anchez\IEEEauthorrefmark{1*}, Enrique Tom\'as Mart\'inez Beltr\'an\IEEEauthorrefmark{1}, Miguel Fern\'andez Llamas\IEEEauthorrefmark{1},\\ G\'er\^ome Bovet\IEEEauthorrefmark{2}, Gregorio Mart\'inez P\'erez\IEEEauthorrefmark{1}, Alberto Huertas Celdr\'an\IEEEauthorrefmark{3}}  


    \IEEEauthorblockA{\IEEEauthorrefmark{1}Department of Information and Communications Engineering, University of Murcia, 30100--Murcia, Spain}
    \IEEEauthorblockA{\IEEEauthorrefmark{2}Cyber-Defence Campus, armasuisse Science \& Technology, CH--3602 Thun, Switzerland}

    \IEEEauthorblockA{\IEEEauthorrefmark{3}Communication Systems Group CSG, Department of Informatics, University of Zurich UZH, CH--8050 Zürich, Switzerland}

    \IEEEauthorblockA{[pedromiguel.sanchez, enriquetomas, m.fernandezllamas, gregorio]@um.es, gerome.bovet@armasuisse.ch, huertas@ifi.uzh.ch}
}

\maketitle

\begin{abstract}

Decentralized Federated Learning (DFL) trains models in a collaborative and privacy-preserving manner while removing model centralization risks and improving communication bottlenecks. However, DFL faces challenges in efficient communication management and model aggregation within decentralized environments, especially with heterogeneous data distributions. Thus, this paper introduces ProFe, a novel communication optimization algorithm for DFL that combines knowledge distillation, prototype learning, and quantization techniques. ProFe utilizes knowledge from large local models to train smaller ones for aggregation, incorporates prototypes to better learn unseen classes, and applies quantization to reduce data transmitted during communication rounds. The performance of ProFe has been validated and compared to the literature by using benchmark datasets like MNIST, CIFAR10, and CIFAR100. Results showed that the proposed algorithm reduces communication costs by up to $\approx$40-50\% while maintaining or improving model performance. In addition, it adds $\approx$20\% training time due to increased complexity, generating a trade-off. 

\end{abstract}


%

\IEEEpeerreviewmaketitle

\section{Introduction}

Over the past few years, Federated Learning (FL) has emerged as a pivotal methodology in Artificial Intelligence, primarily due to its ability to train models without centralizing data \cite{mcmahan2017communication}. This capability is especially critical in scenarios where data privacy and security are paramount. 
Building upon the foundations of FL, Decentralized Federated Learning (DFL) has been proposed as a natural progression of FL, facilitating collaboration among distributed nodes without a central server \cite{beltran2023decentralized}. This evolution aims to address inherent challenges associated with traditional FL, such as communication bottlenecks and risks linked to model centralization.

The interest in DFL is driven by the need for models that can operate efficiently in distributed environments. However, this approach introduces new challenges, including the efficient management of communication between nodes and the aggregation of models in the absence of a central coordinator \cite{liu2022decentralized}. To mitigate these challenges, various techniques have been proposed to optimize communication in DFL by reducing the model size, decreasing the data transmitted between nodes \cite{Wu_2022}.

Recent literature approaches have applied techniques such as Knowledge Distillation (KD), Prototype Learning, quantization, and pruning to minimize communication overhead without significantly compromising model performance. KD enables the creation of smaller, efficient models by transferring knowledge from larger models, reducing both computational and communication demands \cite{gou2021knowledge}. Prototype learning simplifies data representation by modeling essential data characteristics that are common between classes, sharing these representations instead of the complete model \cite{huang2023rethinking}. Finally, quantization reduces the precision of model parameters, while pruning reduces the number of parameters in the network, both leading to smaller model sizes \cite{xu2021accelerating}.

Despite these advancements, several open issues concerning communication optimization in DFL persist. \textit{(i)} A significant challenge is to maintain model performance while reducing size; techniques like KD, prototype learning, and quantization can lead to loss of critical information, adversely affecting model accuracy. \textit{(ii)} The heterogeneity of data distributions (non-IID) and computational capabilities across different nodes further complicates the implementation of these techniques, as most existing methods assume a level of uniformity that is often unrealistic in practical scenarios. 
Therefore, reducing communications while maintaining model performance in DFL is an open challenge that requires additional efforts.

To tackle these issues, the main contribution of this work is ProFe, a novel communication optimization algorithm for DFL that combines KD, prototypes, and quantization techniques. ProFe uses the knowledge acquired by a local large model during each training round to transfer it to a smaller model for aggregation. Then, prototypes are used for the improved learning of unseen classes. Finally, quantization provides an extra optimization in the number of bytes sent during each round. The communication and model performance of ProFe have been evaluated with MNIST, CIFAR10, and CIFAR100 datasets. Additionally, the most representative literature solutions are compared with the proposed approach. The results showed that ProFe effectively reduces communication costs of $\approx$50–60\% while maintaining or even improving model performance, particularly in non-independent and identically distributed settings. Despite introducing an acceptable training time overhead of $\approx$20\%, ProFe outperforms existing methods by balancing communication efficiency and model accuracy across various scenarios.


\section{Related Work}
\label{sec:related}

Several works in the literature have explored the communication optimization problem in FL from various perspectives. In terms of KD, FedKD \cite{Wu_2022} proposes an FL framework where each client stores a larger teacher model and a shared smaller student model, while a central server coordinates the collaborative learning. Federated Mutual Learning (FML) \cite{shen2023federated} is a framework to address data, model, and objective heterogeneity in FL. A "meme model" serves as an intermediary between personalized and global models, enabling knowledge transfer via Deep Mutual Learning on local data. For objective heterogeneity, FML uses a shared global model with selective components, keeping the personalized model task-specific. FML achieves higher accuracy and faster convergence than other FL methods like FedAvg and FedProx, though communication usage is not reported.

From the prototype perspective, FedProto \cite{tan2022fedproto} uses prototype aggregation in a heterogeneous FL framework, transmitting only prototypes between nodes to improve model accuracy and reduce communication costs. This enables varied model architectures across nodes, as long as prototypes remain consistent. However, nodes are limited to learning only labels present in their local data, as models align closely with known label prototypes. Combining prototypes and KD, FedGPD \cite{wu2024global} tackles data heterogeneity in FL by using global class prototypes to guide local training, aligning local objectives with the global optimum. This approach reduces data distribution issues, improving global model performance and achieving 0.22\% to 1.28\% higher accuracy than previous methods.

Using pruning and quantization, FL-PQSU \cite{9366879} introduces techniques that significantly reduce computational and communication costs in FL models with minimal impact on performance. By applying initial model pruning to eliminate unnecessary connections and transmitting only updates that meet specific conditions, unnecessary data transfer is reduced. Additionally, all communication is performed using quantized data types. This combination reduces the model size from 112MB to just 2.83MB (approximately 2.5\% of the original size) with a minimal accuracy loss of 0–1.9\%, and quadruples the communication speed with the server.

\begin{table}[ht!]
\scriptsize
\centering
\caption{Literature Comparison}
\label{tab:related}
\begin{tabular}{P{1.3 cm} P{0.5 cm} P{1.1 cm} P{1.4 cm} P{0.8 cm} P{1.2 cm}}

\textbf{Sol. Year} & \textbf{Arc.} & \textbf{Approach} & \textbf{Dataset} & \textbf{Distrib.} & \textbf{Model} \\
\hline \hline


FL-PQSU \cite{9366879} (2021) & CFL & Prune, Quant. & CIFAR10 & IID & CNN  \\\hline

FedProto \cite{tan2022fedproto} (2022) & CFL & Prototype & MNIST& IID, non-IID  & CNN \\\hline

FedKD \cite{Wu_2022} (2022) & CFL & KD & MIND, SMM4H & non-IID & UniLM-Base \\\hline

FML \cite{shen2023federated} (2023) & CFL & KD & MNIST, CIFAR10/100 & IID, non-IID & LeNet5, CNN, MLP \\\hline

FedGPD \cite{wu2024global} (2024) & CFL & Prototype, KD & MNIST, CIFAR10/100 & non-IID & CNN, Resnet18 \\\hline

\textbf{ProFe} (This work) & DFL & Prototype, KD, Quan. & MNIST, CIFAR10/100 & IID, non-IID & CNN, Resnet8/18 \\\hline

\end{tabular}
\end{table}

\tablename~\ref{tab:related} compares the previous solutions. As can be seen, existing KD-based solutions rely on central coordination and do not consider the particularities of DFL. Prototype-based methods reduce communication costs by sharing class prototypes, but struggle with non-overlapping label distributions, preventing clients from learning unseen classes. Therefore, there is a need for a decentralized approach that combines KD and prototype learning to enable efficient communication, support heterogeneous data and models, and allow clients to learn unseen classes without a central server.

\section{ProFe Algorithm Design}
\label{sec:algorithm}

\begin{figure}[ht!]
    \centering
    \includegraphics[width=\columnwidth,trim={15 15 15 15} ,clip=true]{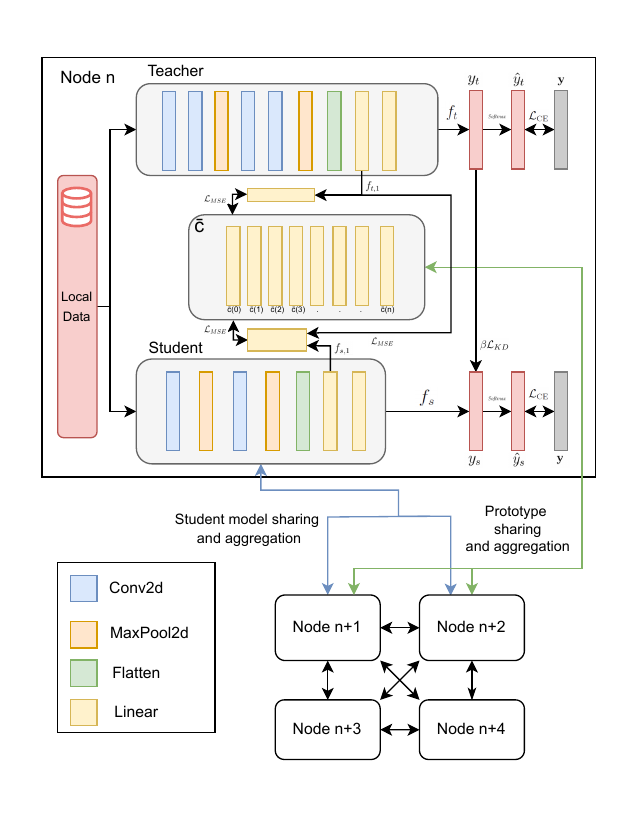}
    \caption{ProFe Algorithm for Model Training in DFL}
    \label{fig:architecture}
\end{figure}

This section presents ProFe, the proposed algorithm for model compression in DFL. \figurename~\ref{fig:architecture} describes the steps taken in each federation node to perform the learning process. Next, the details of KD and prototype learning are given, explaining later the proposed joint approach for model training.

\subsection{Knowledge Distillation}

The implementation of KD between the teacher and student models follows the response-based distillation approach. It begins by calculating \( \mathbf{p}_s \) and \( \mathbf{p}_t \), the smoothed logarithmic probability distributions for the student and teacher models. For the student model, a \textit{Log-Softmax} function is used, while for the teacher model, a standard \textit{Softmax} function is applied.

Let \( n \) be the number of output layers in the neural network, and let \(\mathbf{p}_s = (p_{s,1}, p_{s,2}, \ldots, p_{s,n})\) and \(\mathbf{p}_t = (p_{t,1}, p_{t,2}, \ldots, p_{t,n})\) represent the smoothed probability distributions for the student and teacher models, respectively. Then,
\[
p_{s,j} = \ln \left( \frac{e^{y_{s,j} / T}}{\sum_{k=1}^{n} e^{y_{s,k} / T}} \right) \quad \text{and} \quad p_{t,j} = \frac{e^{y_{t,j} / T}}{\sum_{k=1}^{n} e^{y_{t,k} / T}},
\]

where \(y_s = (y_{s,1}, y_{s,2}, \ldots, y_{s,n})\) and \(y_t = (y_{t,1}, y_{t,2}, \ldots, y_{t,n})\) are the outputs of the student and teacher networks, respectively, and \(T\) is the \textit{temperature} parameter in KD that smooths the probability distributions.

The Kullback-Leibler divergence is then used to quantify how the probability distribution \( \mathbf{p}_t \) deviates from \( \mathbf{p}_s \) as:
\[
\text{KL}(\mathbf{p}_t || \mathbf{p}_s) = \sum_{j=1}^{n} p_{t,j} \ln \left( \frac{p_{t,j}}{p_{s,j}} \right).
\]

The total KD loss with temperature is calculated as:
\[
\mathcal{L}_{KD}(y_s, y_t) = \text{KL}(\mathbf{p}_t || \mathbf{p}_s) \times T^2.
\]

In addition, a cross-entropy loss function is incorporated. Let \( y_s = (y_{s,1}, y_{s,2}, \ldots, y_{s,n}) \) be the \textit{Logits} from the student model for \( n \) and \( \mathbf{y} = (y_1, y_2, \ldots, y_n) \) be the vector of true labels in one-hot format. The Softmax function is applied to convert the model outputs into a probability distribution:
\[
\mathbf{q}_s = (q_{s,1}, q_{s,2}, \ldots, q_{s,n}) \quad \text{where} \quad q_{s,j} = \frac{e^{y_{s,j}}}{\sum_{k=1}^{n} e^{y_{s,k}}}.
\]

The cross-entropy loss is then calculated as:
\begin{equation}
\label{eq:LCE}
\mathcal{L}_{\text{CE}}(y_s, \textbf{y}) = -\sum_{j=1}^{n} y_j \ln(q_{s,j}).
\end{equation}

The final loss function of the model when KD is applied is computed as:
\begin{equation}
\label{eq:loss_kd}
\mathcal{L} = \mathcal{L}_{\text{CE}} + \beta \mathcal{L}_{KD},    
\end{equation}

where $\beta$ is a parameter in the interval $[0,1]$ that determines the weight of the distillation loss during training.

\subsubsection{Professor Importance Decay}

In this implementation, $\beta$ is decremented to half of its initial value after each federated round. A new variable, $\beta_{\text{limit}}$, is introduced so that when $\beta$ falls below this threshold, it is set to zero. This adjustment allows the student to bypass training the teacher model, thereby saving computational time, energy, and memory. Moreover, the use of the previously mentioned functions enhances the versatility of the models and simplifies the implementation of communication changes.

\subsection{Prototype Generation}

Implementing the prototype basis in the FL environment will follow the approach presented in FedProto \cite{tan2022fedproto}, but using cross-entropy loss ($\mathcal{L}_{CE}$) instead of negative-log likehood loss ($\mathcal{L}_{NLL}$). This approach simplifies the classification process and improves results, particularly when data is extremely limited. Moreover, it only shares class prototypes, significantly reducing communication costs.

Let $\mathbb{C}$ be the set of all existing labels, and $\mathbb{C}_i$ the subset of labels known by node $i$. The prototype $C^{(j)}_i$ , representing the $j$-th class in 
$\mathbb{C}_i$, is defined as the average of the vector representations of elements in class $j$. These representations are the output from the model inference process, taken prior to the final predictions. As shown in \figurename~\ref{fig:architecture}, prototypes are calculated using the output of the model first linear layer.

The inference function at node $i$ is expressed as 
$f_i = f_{i,2} \circ f_{i,1}$, where $f_{i,1}$ provides the vector representations used to calculate prototypes, and 
$f_{i,2}$ computes the final model predictions using the representations in a standard manner. The prototype $C^{(j)}$ for node $i$ is computed as: 
\begin{equation} \label{eq:proto
} C^{(j)}i = \frac{1}{|D{i,j}|} \sum_{(x,y) \in D_{i,j}}f_{i,1}(x), \end{equation}

where $D_{i,j}$ is the subset of the local dataset $D_i$, and all data points belong to class $j$.

After each round, prototypes are calculated and sent to other nodes. Each client receives the prototypes from other nodes, allowing them to compute global prototypes 
$\overline{C}^{(j)}$ for each label. These global prototypes are defined as: 

\begin{equation} \label{eq:globalproto
} \overline{C}^{(j)} = \frac{1}{|\mathcal{N}j|} \sum_{i \in \mathcal{N}j} \frac{|D{i,j}|}{N_j} C^{(j)}_i, \end{equation}

where $C^{(j)}_i$ denotes the prototype for class $j$ at node $i$, $\mathcal{N}_j$ represents the set of clients that know class  $j$, and $N_j$ is the number of instances of class $j$ across all nodes.

Once the global prototypes are defined, the inference function at node $i$ no longer uses $f_{i,2}$. Instead, the predicted label $\hat{y}$ is determined by: 
\begin{equation} \label{eq:dist_proto} \hat{y} = \arg \min_j \left\| f_{i,1}(x) - \overline{C}^{(j)} \right\|_2, \end{equation} 
where $\left\| \cdot \right\|_2$ is the Euclidean distance. The predicted class will be the label of the prototype closest to the input vector representation.

The loss function of the models is computed as the weighted sum of two loss components. First, cross-entropy ($\mathcal{L}_{CE}$) is used, as defined in the previous section (see Equation \ref{eq:LCE}). The second component is the mean squared error loss ($\mathcal{L}_{MSE}$), calculated between the current output's vector representations ($f_{i,1}(x)$) and the global prototype of class 
$j$, which corresponds to the true label of the input $x$. This is given by:
\begin{equation} \label{eq:loss_proto}
\mathcal{L}_{MSE}\left(f_{i,1}(x), \overline{C}^{(j)}\right) = \frac{1}{n} \sum_{k=1}^{n} \left(f_{i,1,k}(x) - \overline{C}^{(j)}_k\right)^2,\end{equation}

\noindent where \(n\) is the number of elements in the vector representations, \(f_{i,1,k}(x)\) is the \( k \)-th element of the vector representation \( f_{i,1}(x) \), and \(\overline{C}^{(j)}_k\)is the \( k \)-th element of the global prototype for class \( j \). All these elements are tensors of the same dimension, ensuring the feasibility of the operation. Thus, the final loss function is calculated as:
\begin{equation}
    \mathcal{L} = \mathcal{L}_{CE} + \beta\mathcal{L}_{MSE},\end{equation}
    
\noindent where in this case $\beta$ is set to 1 \cite{tan2022fedproto}.

\subsection{Knowledge Distillation and Prototype Integration}

Once the separate processes of KD and Prototype generation have been described separately, they are integrated together in the proposed approach (see \figurename~\ref{fig:architecture}). This perspectives ensures achieving the benefits of both techniques, improving the performance of applying them independently.

In this approach, student model is trained using the cross-entropy loss function ($\mathcal{L}_{CE}$), the loss between global prototypes and the student's intermediate vector representations ($\mathcal{L}_{MSE}$), the KD loss between the logits of the teacher and student models ($\mathcal{L}_{KD}$), and an additional loss between the intermediate vector representations of the student and teacher models ($\mathcal{L}_{MSE}$). This last loss employs the same function as the loss between the global prototypes and the student vector representations but is applied with different parameters, thus requiring no further development. 

The teacher model is trained using the cross-entropy loss function ($\mathcal{L}_{CE}$), the loss between global prototypes, the teacher’s intermediate vector representations ($\mathcal{L}_{MSE}$).

The student model overall loss is expressed as:
\begin{equation}
\begin{split}
\mathcal{L}_s = \mathcal{L}_{CE}(y_s, \textbf{y}) + \beta_s\mathcal{L}_{MSE}(f_{s,1}(x),\overline{C}^{(j)}) + \\\alpha_s[\mathcal{L}_{KD}(y_s,y_t) +\mathcal{L}_{MSE}(f_{s,1}(x), f_{t,1}(x))].
\end{split}
\end{equation}

While the teacher model loss is given by:
\begin{equation}
\mathcal{L}_t = \mathcal{L}_{CE}(y_t, \textbf{y}) + \beta_t\mathcal{L}_{MSE}(f_{t,1}(x),\overline{C}^{(j)}). 
\end{equation}

Where $j$ is the true label corresponding to input $x$. The notation is the same used throughout this section, where \textbf{y} represents the one-hot encoded vector, where the correct class at position $j$ is set to one. $y_s$ and $y_t$ denote the logits of the student and teacher models. 
$f_{s,1}(x)$ and $f_{t,1}(x)$ are the intermediate vector representations of the student and teacher models, and $\overline{C}^{(j)}$ is the global prototype for label $j$.

To compute all these loss functions, the inference methods for student and teacher models must return the logits, intermediate vector representations, and convolutional layers. To achieve this, a dedicated inference function is defined for training purposes in both the student and teacher models.

\subsection{Quantization}

To reduce communication costs in model and prototype exchanges, 16-bit quantization is applied, halving the data size transferred. The quantization function is defined as:

\[
Q(x) = \left\lfloor \frac{x}{\Delta} + 0.5 \right\rfloor \Delta,
\]

where $x$ is the model parameter, $Q(x)$ is the quantized value, and $\Delta$ is the quantization step. In this case, $\Delta$ is chosen to fit the values into the 16-bit range.

When models and prototypes from other nodes arrive for aggregation, inverse quantization approximates the original values using:
\[ x' = Q(x) \cdot \Delta. \]
After inverse quantization, model training continues at 32-bit precision, preserving performance while reducing communication costs by half. The precision loss from 16-bit quantization is minimal, ensuring accuracy during training and aggregation.

\begin{figure*}[htpb!]
    \centering
    \includegraphics[width=\textwidth]{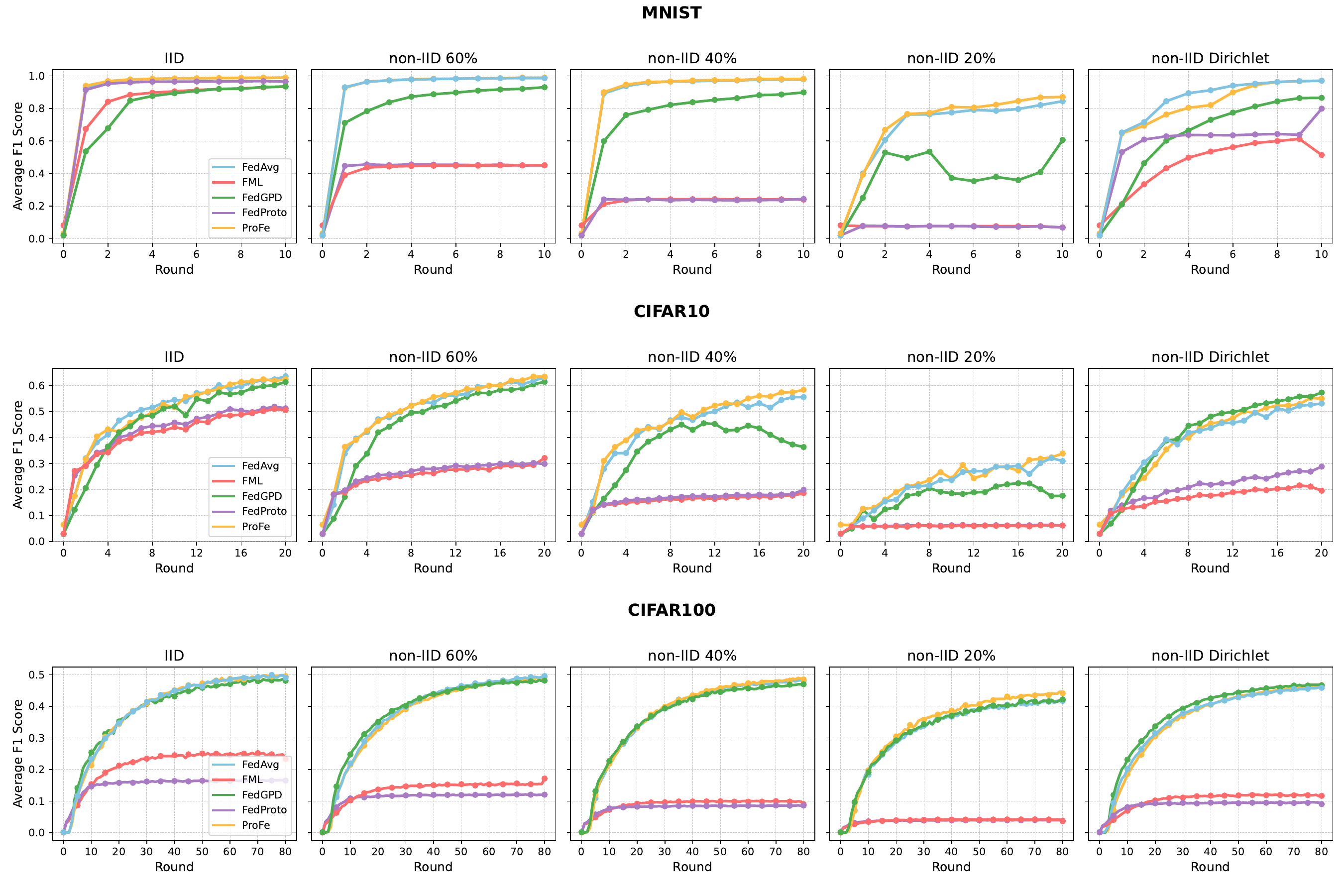}
    \caption{F1-Score Comparison Between the Proposed Algorithm (ProFe) and the Literature}
    \label{fig:validation}
\end{figure*}

\section{Experiments}
\label{sec:validation}

A pool of experiments has evaluated the communication cost, model performance, and training time of ProFe. In addition it has been compared to FedAvg, FedGPD \cite{wu2024global}, FML \cite{shen2023federated}, and FedProto \cite{tan2022fedproto}, the most relevant works of the literature. For the experiments, MNIST, CIFAR10, and CIFAR100 datasets (three well-known benchmark datasets focused on image classification) have been considered. To test data diversity, five different configurations are tested: IID, non-IID with 60\% classes present in each client, non-IID with 40\% classes, non-IID with 20\% classes, and non-IID Dirichlet with $\alpha=0.5$. Before data splitting, 10\% of each dataset is selected as a global test dataset for evaluation. Then, in each node, the local dataset is divided into 80\%/20\% for training and tests. 

The experiments comprise 20 nodes following a fully connected topology. For MNIST experiments, 10 rounds of one epoch are executed in the training configuration; for CIFAR10, 20 rounds of three epochs are executed; and for CIFAR100, due to the more complex task, 80 rounds of three epochs are executed. Regarding model setup, for MNIST, a two-layer CNN is chosen as the teacher network, having half of the channels in the student network. For CIFAR10, Resnet18 is chosen as teacher and Resnet8 as student. Finally, for CIFAR100, Resnet32 is chosen as teacher and Resnet18 as student. None of these networks are pre-trained. All the experiments are run on a server with an Intel i7-10700F CPU, two RTX 3080 GPUs, and 98GB RAM. While running the experiments, no other tasks were executed to avoid performance conflicts. For validation, the Nebula framework \cite{beltran2024fedstellar} was used, which includes real communications between nodes using Dockers. The implementation and experiments code is available at \cite{code}.

\subsection{Performance}

The first critical point to validate regarding the proposed approach is the performance comparison with the baseline (FedAvg) and the literature methods. \figurename~\ref{fig:validation} illustrates the average node F1-Score in each training round for the different datasets tested. ProFe performance is aligned with the baseline in all the scenarios, even improving it is many non-IID ones. It achieves $+95\%$ F1-Score for MNIST in IID and most non-IID setups, only degrading to $+80\%$ in non-IID with 20\% classes known. Similarly, it achieves $\approx60\%$ on CIFAR10, only degraded to $+30\%$ on non-IID 20\%. Finally, it achieves $\approx50\%$ on CIFAR100, degraded to $\approx45\%$ on non-IID 20\%.

FedProto works well on the IID case for MNIST. However, when there is a non-IID distribution or the classification task becomes more complex (CIFAR10/100), its performance greatly degrades. A similar situation is observed for FML, which performs close to FedProto in most experiments. ProFe improves both FedProto and FML in all the conducted experiments. FedGPD maintains a performance closer to the baseline and ProFe in most of the experiments, specially in CIFAR100 classification. In contrast, FedGPD performance degrades in the non-IID cases for the rest of the datasets, only achieving the best performance on non-IID Dirichlet for CIFAR10/100, and being surpassed by ProFe in the rest of the cases.

From these results, it can be considered that ProFe is a viable option for DFL scenarios, having a performance aligned with the baseline and improving other methods in most of the cases, specially in non-IID setups. Then, the next vital step is to verify and compare the communication improvement achieved by the proposed solution.

\begin{table*}[htpb!]
    \centering
    \scriptsize
    \caption{Network Bytes Received and Sent (GB) for Different Solutions under IID Distribution (\% to FedAvg)}
    \begin{tabular}{@{}lcccccc@{}}
        \toprule
        \multirow{2}{*}{} & \multicolumn{3}{c}{Bytes Received} & \multicolumn{3}{c}{Bytes Sent} \\ 
        \cmidrule(lr){2-4} \cmidrule(lr){5-7}
        & MNIST & CIFAR10 & CIFAR100 & MNIST & CIFAR10 & CIFAR100 \\ 
        \midrule
        FedAvg & 105.140$\pm$2.376 & 150.713$\pm$0.432 & 1311.300$\pm$0.695 & 105.126$\pm$2.376 & 149.992$\pm$0.432 & 1309.958$\pm$0.695 \\
        FedGPD & 109.736$\pm$1.937 {\color{purple}($\uparrow$ 4\%)} & 148.786$\pm$0.130 {\color{teal}($\downarrow$ 2\%)} & 1325.708$\pm$0.021 {\color{purple}($\uparrow$ 1\%)} & 109.731$\pm$1.935 {\color{purple}($\uparrow$ 4\%)} & 148.624$\pm$0.130 {\color{purple}($\uparrow$ 2\%)} & 1322.912$\pm$0.021 {\color{purple}($\uparrow$ 1\%)} \\
        FML  & 123.994$\pm$2.254 {\color{purple}($\uparrow$ 18\%)} & 157.562$\pm$0.202 {\color{purple}($\uparrow$ 4\%)} & 1325.497$\pm$0.485 {\color{purple}($\uparrow$ 1\%)} & 123.933$\pm$2.254 {\color{purple}($\uparrow$ 18\%)} & 156.861$\pm$0.202 {\color{purple}($\uparrow$ 4\%)} & 1324.766$\pm$0.485 {\color{purple}($\uparrow$ 1\%)} \\
        FedProto & 8.761$\pm$0.002 {\color{teal}($\downarrow$ 91\%)} & 7.508$\pm$0.036 {\color{teal}($\downarrow$ 95\%)} & 21.218$\pm$0.001 {\color{teal}($\downarrow$ 98\%)} & 8.760$\pm$0.002 {\color{teal}($\downarrow$ 91\%)} & 7.341$\pm$0.027 {\color{teal}($\downarrow$ 95\%)} & 21.222$\pm$0.001 {\color{teal}($\downarrow$ 98\%)} \\
        \textbf{ProFe} & 62.404$\pm$1.653 {\color{teal}($\downarrow$ 41\%)} & 77.299$\pm$0.004 {\color{teal}($\downarrow$ 49\%)} & 677.61$\pm$0.008 {\color{teal}($\downarrow$ 49\%)} & 62.403$\pm$1.653 {\color{teal}($\downarrow$ 41\%)} & 77.302$\pm$0.004 {\color{teal}($\downarrow$ 49\%)} & 674.734$\pm$0.008 {\color{teal}($\downarrow$ 49\%)} \\
        \bottomrule
    \end{tabular}
    \label{tab:network_bytes}
\end{table*}

\subsection{Communication and Training Time Costs}

Another key aspect to consider and optimize is the network cost of all the communications, which exponentially explodes in DFL scenarios. \tablename~\ref{tab:network_bytes} shows the bytes sent and received on average by each node in the experiments. Only the network usage in the IID configuration is shown for space purposes, as the local node data distribution does not vary the network usage proportion between methods.

It can be seen how FedProto \cite{tan2022fedproto} is the best solution in all cases in terms of communications. In the second place, ProFe reduces $\approx$40-50\% of the communication cost from the baseline FedAvg and the other solutions compared, which have even more communication cost than the baseline. This improvement is particularly relevant in DFL, where communication overhead is a critical bottleneck affecting the scalability and practicality of deployment in real-world applications.

In terms of training time, \tablename~\ref{tab:elapsed_time} shows the average time required by the nodes to complete the experiments using each dataset. It can be seen how FedProto is the clear winner in all the datasets. However, it has been shown that its performance is greatly degraded when the classification tasks get more complex. Other literature methods (FedGPD and FML) are close to the base FedAvg, adding between 5-11\% to the base time. In contrast, the proposed method has a larger training time due to its increased complexity, adding around 18-20\% in the case of CIFAR10 and CIFAR100. In contrast, the time is almost identical to FedAvg in the case of MNIST, where simpler networks are used. This increase in the training time is natural due to the combination of KD and prototypes in the proposed technique, being acceptable in scenarios where optimizing communications is a priority in exchange for sacrificing a proportion of training time.

\begin{table}[htpb]
    \centering
    \scriptsize
    \caption{Elapsed Time (seconds) under IID Distribution (\% to FedAvg)}
    \resizebox{\columnwidth}{!}{
    \begin{tabular}{@{}lccc@{}}
        \toprule
        & MNIST & CIFAR10 & CIFAR100 \\ 
        \midrule
        FedAvg & 887.3$\pm$44.24 & 1962.95$\pm$27.40 & 17046.7$\pm$90.75 \\
        FedGPD & 1047.25$\pm$3.56 {\color{purple}($\uparrow$ 18\%)} & 2185.0$\pm$39.18 {\color{purple}($\uparrow$ 11\%)} & 18449.2$\pm$91.83 {\color{purple}($\uparrow$ 8\%)} \\
        FML & 933.7$\pm$44.19 {\color{purple}($\uparrow$ 5\%)} & 2046.65$\pm$34.65 {\color{purple}($\uparrow$ 4\%)} & 18005.35$\pm$111.03 {\color{purple}($\uparrow$ 6\%)} \\
        FedProto & 307.75$\pm$0.94 {\color{teal}($\downarrow$ 65\%)} & 554.15$\pm$1.15 {\color{teal}($\downarrow$ 72\%)} & 4213.7$\pm$0.64 {\color{teal}($\downarrow$ 75\%)} \\
        \textbf{ProFe} & 878.4$\pm$3.74 {\color{teal}($\downarrow$ 1\%)} & 2342.1$\pm$8.54 {\color{purple}($\uparrow$ 19\%)} & 20209.05$\pm$84.96 {\color{purple}($\uparrow$ 19\%)} \\
        \bottomrule
    \end{tabular}}
    \label{tab:elapsed_time}
\end{table}

In conclusion, ProFe offers a $\approx$40-50\% communication reduction and an increase of $\approx$20\% in the training time. While this trade-off is reasonable for most scenarios, it is a matter of each use case to decide which approach is the most suitable.

\section{Conclusion and Future Work}
\label{sec:conclusion}


This paper introduced ProFe, a novel communication optimization algorithm for DFL that integrates KD, prototype learning, and quantization techniques to reduce communication overhead without sacrificing model accuracy. Experiments with benchmark datasets such as MNIST, CIFAR10, and CIFAR100 demonstrated that ProFe effectively reduces communication costs by approximately 40-50\% compared to the baseline and other leading methods. Importantly, ProFe maintained or even improved model performance in most cases, particularly in non-IID data setups where data heterogeneity among nodes is significant. 
In contrast, the integration of these techniques increases training time by approximately 18-20\%. However, this trade-off is acceptable in scenarios where communication bandwidth is a critical constraint. Future work aims to reduce ProFe computational complexity through techniques like model pruning and adaptive optimization algorithms. Incorporating methods such as transfer learning may enable ProFe to quickly adapt to new tasks. 


\section*{Acknowledgements}
This work was supported by \textit{(a)} the Swiss Federal Office for Defense Procurement (armasuisse) with DATRIS (CYD-C-2020003) project, \textit{(b)} University of Zürich UZH, \textit{(c)} 21629/FPI/21, Fundación Séneca, Región de Murcia (Spain), and \textit{(d)} strategic project DEFENDER from the Spanish National Institute of Cybersecurity (INCIBE) and by the Recovery, Transformation and Resilience Plan, Next Generation EU.

\bibliographystyle{IEEEtran}  
\bibliography{references}

\begin{thebibliography}{10}
\providecommand{\url}[1]{#1}
\csname url@samestyle\endcsname
\providecommand{\newblock}{\relax}
\providecommand{\bibinfo}[2]{#2}
\providecommand{\BIBentrySTDinterwordspacing}{\spaceskip=0pt\relax}
\providecommand{\BIBentryALTinterwordstretchfactor}{4}
\providecommand{\BIBentryALTinterwordspacing}{\spaceskip=\fontdimen2\font plus
\BIBentryALTinterwordstretchfactor\fontdimen3\font minus \fontdimen4\font\relax}
\providecommand{\BIBforeignlanguage}[2]{{%
\expandafter\ifx\csname l@#1\endcsname\relax
\typeout{** WARNING: IEEEtran.bst: No hyphenation pattern has been}%
\typeout{** loaded for the language `#1'. Using the pattern for}%
\typeout{** the default language instead.}%
\else
\language=\csname l@#1\endcsname
\fi
#2}}
\providecommand{\BIBdecl}{\relax}
\BIBdecl

\bibitem{mcmahan2017communication}
B.~McMahan, E.~Moore, D.~Ramage, S.~Hampson, and B.~A. Arcas, ``Communication-efficient learning of deep networks from decentralized data,'' in \emph{Artificial intelligence and statistics}.\hskip 1em plus 0.5em minus 0.4em\relax PMLR, 2017, pp. 1273--1282.

\bibitem{beltran2023decentralized}
E.~T.~M. Beltr{\'a}n \emph{et~al.}, ``Decentralized federated learning: Fundamentals, state of the art, frameworks, trends, and challenges,'' \emph{IEEE COMST}, 2023.

\bibitem{liu2022decentralized}
W.~Liu, L.~Chen, and W.~Zhang, ``Decentralized federated learning: Balancing communication and computing costs,'' \emph{IEEE Transactions on Signal and Information Processing over Networks}, vol.~8, pp. 131--143, 2022.

\bibitem{Wu_2022}
C.~Wu, F.~Wu, L.~Lyu, Y.~Huang, and X.~Xie, ``Communication-efficient federated learning via knowledge distillation,'' \emph{Nature Communications}, vol.~13, 2022.

\bibitem{gou2021knowledge}
J.~Gou, B.~Yu, S.~J. Maybank, and D.~Tao, ``Knowledge distillation: A survey,'' \emph{International Journal of Computer Vision}, vol. 129, no.~6, pp. 1789--1819, 2021.

\bibitem{huang2023rethinking}
W.~Huang, M.~Ye, Z.~Shi, H.~Li, and B.~Du, ``Rethinking federated learning with domain shift: A prototype view,'' in \emph{IEEE/CVF Conference on Computer Vision and Pattern Recognition (CVPR)}, 2023, pp. 16\,312--16\,322.

\bibitem{xu2021accelerating}
W.~Xu, W.~Fang, Y.~Ding, M.~Zou, and N.~Xiong, ``Accelerating federated learning for iot in big data analytics with pruning, quantization and selective updating,'' \emph{IEEE Access}, vol.~9, pp. 38\,457--38\,466, 2021.

\bibitem{shen2023federated}
T.~Shen \emph{et~al.}, ``Federated mutual learning: a collaborative machine learning method for heterogeneous data, models, and objectives,'' \emph{Frontiers of Information Technology \& Electronic Engineering}, vol.~24, no.~10, pp. 1390--1402, 2023.

\bibitem{tan2022fedproto}
Y.~Tan \emph{et~al.}, ``Fedproto: Federated prototype learning across heterogeneous clients,'' in \emph{Proceedings of the AAAI Conference on Artificial Intelligence}, vol.~36, no.~8, 2022, pp. 8432--8440.

\bibitem{wu2024global}
S.~Wu \emph{et~al.}, ``Global prototype distillation for heterogeneous federated learning,'' \emph{Scientific Reports}, vol.~14, no.~1, p. 12057, 2024.

\bibitem{9366879}
W.~Xu, W.~Fang, Y.~Ding, M.~Zou, and N.~Xiong, ``Accelerating federated learning for iot in big data analytics with pruning, quantization and selective updating,'' \emph{IEEE Access}, vol.~9, pp. 38\,457--38\,466, 2021.

\bibitem{beltran2024fedstellar}
E.~T.~M. Beltr{\'a}n \emph{et~al.}, ``Fedstellar: A platform for decentralized federated learning,'' \emph{Expert Systems with Applications}, vol. 242, p. 122861, 2024.

\bibitem{code}
M.~Fern{\'a}ndez~Llamas \emph{et~al.}, ``{NEBULA: A Platform for Decentralized Federated Learning - Communication Optimization},'' \url{https://github.com/CyberDataLab/nebula/tree/feature/communication_optimization}, 2024, [Online; accessed 30-October-2024].

\end{thebibliography}

\end{document}